\documentclass{article}
\usepackage{ijcai22}

\usepackage{times}
\usepackage{soul}
\usepackage{url}
\usepackage[hidelinks]{hyperref}
\usepackage[utf8]{inputenc}
\usepackage[small]{caption}
\usepackage{graphicx}
\usepackage{amsmath}
\usepackage{amsthm}
\usepackage{booktabs}
\usepackage{algorithm}
\usepackage{algorithmic}
\usepackage{bigstrut}
\usepackage{multirow}
\usepackage{multirow}
\usepackage{bigstrut}
\usepackage{graphicx}
\usepackage{longtable}
\usepackage{subfigure}
\usepackage{diagbox}
\title{AdvSmo: Black-box Adversarial Attack by Smoothing Linear Structure of Texture}

\author{
Hui Xia\and
Rui Zhang\and
Shuliang Jiang\and
Zi Kang
\affiliations
College of Computer Science and Technology, Ocean University of China, Qingdao 266100, China\\
\emails
\{zhang\_rui0504)\}@163.com
}

\begin{document}

\maketitle

\begin{abstract}
Black-box attacks usually face two problems: poor transferability and the inability to evade the adversarial defense. To overcome these shortcomings, we create an original approach to generate adversarial examples by smoothing the linear structure of the texture in the benign image, called AdvSmo. We construct the adversarial examples without relying on any internal information to the target model and design the imperceptible-high attack success rate constraint to guide the Gabor filter to select appropriate angles and scales to smooth the linear texture from the input images to generate adversarial examples. Benefiting from the above design concept, AdvSmo will generate adversarial examples with strong transferability and solid evasiveness. Finally, compared to the four advanced black-box adversarial attack methods, for the eight target models, the results show that AdvSmo improves the average attack success rate by 9\% on the CIFAR-10 and 16\% on the Tiny-ImageNet dataset compared to the best of these attack methods.
\end{abstract}

\section{Introduction}
Deep learning (DL), with its powerful characterization ability and generalization performance, has received attention from many researchers in academia and industry and has been widely used in image classification, target detection, machine translation, and other fields. Google applies deep learning to speech recognition and image recognition, Microsoft leverages DL for machine translation, and Netflix and Amazon use DL to understand customer behavior habits. Although deep learning has made extraordinary breakthroughs in several fields, it does not mean that DL has matured. And its security problems caused by its lack of robustness and vulnerability to adversarial attacks are still a major challenge today. For example, a simple noise added to a photo of a person's face can trick the face recognition system into identifying a stranger as a familiar person. Worse still, the transferability of the adversarial attack poses a more serious security risk to the model, even if the attacker does not know what the model is able to attack successfully. It is clear that effective defense against adversarial attacks is the key to promoting the widespread use of deep learning.

\begin{figure}[t]
    \centering
    \includegraphics[scale=0.22]{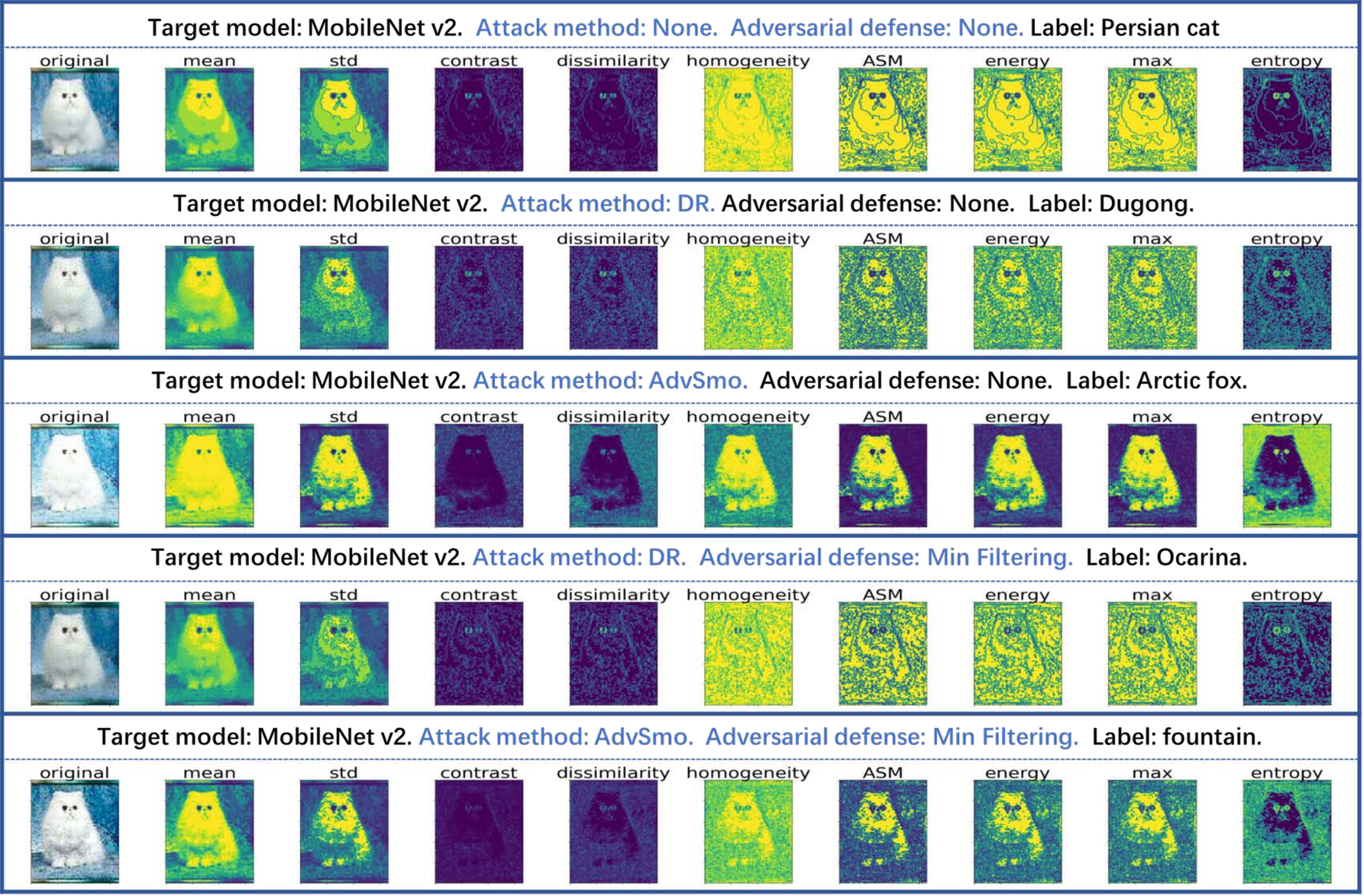}
    \caption{The image's texture (labeled `original') is extracted by using Gray Level Co-occurrence Matrix. Although the adversarial examples generated by DR and AdvSmo are both evasive, the attack success rate, transferability.}
\end{figure}
Adversarial attacks are mainly categorized into three types: white-box attacks, gray-box attacks, and black-box attacks. The white-box attack means that the adversary is free to use any information of the target model (e.g., network structure and model parameters) to construct an adversarial example. But this implies that the adversary needs to thoroughly understand all the information of the target model before launching the attack, which is not feasible in real-world application scenarios. Gray-box attacks are between white-box and black-box attacks, and only some information about the target model, such as the output probability of the model and the decision boundary of the model, is needed to launch the attack, but this partial information is still difficult to obtain in practical applications. Black-box attacks do not need to access the internal structure information of the target model and design effective perturbations to add them to the input images to generate adversarial examples only by accessing the inputs and outputs of the target model. However, such attacks do not directly access the internal structure information of the target model, but they use the structure and parameters of the substitution model to construct the adversarial examples, resulting in poor transferability of adversarial examples.

To solve the above problems, we propose a black-box adversarial attack from the perspective of image processing by smoothing the linear structure of the texture to construct the adversarial examples. We perform a comprehensive evaluation of the proposed scheme. It achieves high attack success rates against all eight target models on CIFAR 10 and Tiny-ImageNet datasets, as shown in Figure 1. The main contributions are as follows:

\begin{itemize}
 \item  To the best of our knowledge, this is the first black-box adversarial attack scheme by smoothing the linear structure of texture with Gabor filtering. Since the scheme smoothes all linear textures in the benign image, the generated adversarial examples have a high attack success rate. They do not rely on the details of the target model to design the adversarial perturbation, so the generated adversarial examples are highly transferable. Furthermore, the adversarial defense technique can only recover a small number of the linear texture of the adversarial example, so the scheme has a solid evasion for adversarial defenses.

 \item We verify the validity of AdvSmo by comparing the attack success rates of four black box attack schemes against eight threat models. Specifically, for the eight threat models, the average attack success rate of AdvSmo is improved by 37\% compared to DR, 47\% compared to BIA, 13\% compared to GUA, and 51\% compared to NPAttack on CIFAR-10 and Tiny-ImageNet dataset.

\end{itemize}

\section{Related work}
This section focuses on two types of black-box attack schemes, query-based and transferability-based.

\subsection{Query-based black-box attack}
Query-based attacks can be subdivided into score-based and decision-based attacks according to the information about the target model used by the attacker to generate the adversarial examples. The score-based attack uses the predicted score of the target model and then generates adversarial examples based on the estimated gradient. Meunier \emph{et al.}~\cite{meunier2019yet} proposed an effective discrete alternative method to reduce the huge query cost required to estimate the gradient. Du \emph{et al.}~\cite{du2019query} used meta-learning to estimate the gradient of the target model to reduce the number of queries. However, such attacks inevitably have huge query overhead. Defense mechanisms easily detect adversaries with frequent visits to the model. Decision-based attacks generate adversarial examples based on the output labels of the target model. Maho \emph{et al.} ~\cite{maho2021surfree} proposed a geometric method to generate effective adversarial examples. Chen \emph{et al.}~\cite{chen2020rays} proposed a ray search method for should-label adversarial attacks that greatly reduces the number of queries to the target model. Bai \emph{et al.}~\cite{bai2020improving} proposed a black-box adversarial attack based on neural processes to improve the query efficiency and explored the distribution of adversarial examples around benign inputs with the help of the image structure information characterized by neural processes. Although such attacks can reduce the number of queries of the model, they cannot generate the adversarial example with a high attack success rate. It is still a very challenging task to guarantee a high attack success rate while reducing the number of queries.

\subsection{Transferability-based black-box attack}

Transferability-based attacks generate adversarial examples for the substitution model and then use them to attack the target model. Papernot \emph{et al.}~\cite{papernot2017practical} used the idea of model stealing to train a substitution model with the synthetic dataset labeled by the target model. Ma \emph{et al.}~\cite{ma2021simulating} reduced the number of queries while minimizing the differences with the target model by training a simulator of the target model based on query sequences from multiple existing attack methods. However, transferability-based attacks require the training of substitution models, which undoubtedly increases the cost of the attack significantly, and the information related to the deployed models is unlikely to be compromised. Zhang \emph{et al.}~\cite{zhang2022beyond} build a more practical black-box threat model to overcome this limitation, proposing a beyond ImageNet attack (BIA) to study transferability to black-box domains. Lu \emph{et al.}~\cite{lu2020enhancing} studied the transferability of adversarial examples in the real world and proposed an attack that minimizes the `dispersion' of the internal feature map, overcoming the shortcomings of existing attacks that require task-specific loss functions and/or detection target models. However, the transferability of the above scheme to generate adversarial examples is still relatively poor, and it cannot evade the adversarial defense mechanism.

\begin{figure}[t]
    \centering
    \includegraphics[scale=0.22]{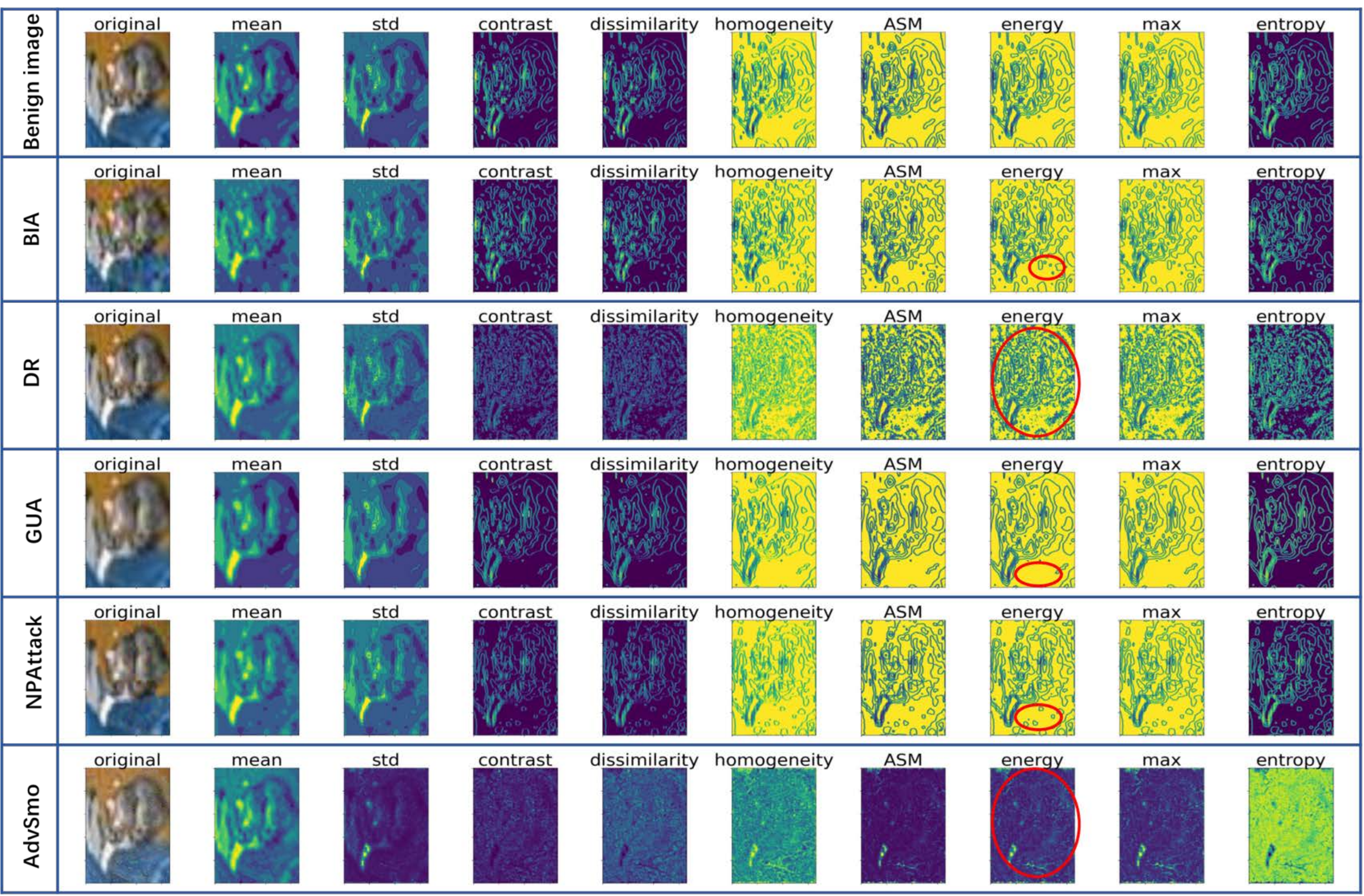}
    \caption{Adversarial example show}
\end{figure}

\section{System Model}

\subsection{Problem Definition}
We try to change the decision features of the Deep Neural Network (DNN) in the input image to generate adversarial examples without affecting human visual perception.
Inspired by Gerihos et al. \cite{geirhos2018imagenet}, we use the Gabor filter to smooth the linear structure of texture in benign images and generate adversarial examples to fool the target classifier. The objective function is

\begin{equation}
  \begin{array}{l}
 \arg \mathop {\max }\limits_c f({X_{adv}},\tilde \vartheta ) = \hat y, \\
 s.t.,{\rm{ }}d{\rm{(}}{X_{adv}},X{\rm{)}} < \eta ,\hat y \ne y, \\
 \end{array}
\end{equation}

$f({X_{adv}},\tilde \vartheta )$ is the target classifier. \emph{X} is the benign image. \emph{R} is the linear of texture that is smoothed from the benign image by using the Gabor filter. ${X_{adv}}$ is the adversarial example, ${X_{adv}} = X - R$. $\eta$ limits the magnitude of smoothed features. $\tilde \vartheta$ is the parameters of the classifier. $\hat y$ is the labels misclassified by the target classifier. \emph{y }is the original label of the benign image. $d( \cdot )$ is the distance measurement.

\subsection{Constructing the Initial Adversarial Example Set}


\begin{equation}
g({k_1},{k_2}) = \exp ( - {{{k_1}^{{{'}}2} + {\gamma ^2}{k_2}^{'2}} \over {2{\sigma ^2}}})\exp (i(2\pi {{{k_1}^{'}} \over \lambda } + \psi ))
\end{equation}

\begin{equation}
{k_1}^{{'}} = {k_1}\cos \theta  + {k_2}\sin \theta
\end{equation}

\begin{equation}
{k_2}^{'} =  - {k_1}\sin \theta  + {k_2}\cos \theta
\end{equation}

 $\lambda$ is the wavelength. $\psi$ is the phase of the Cosine function. $\gamma$ is the spatial aspect ratio. ${k_1}$ and ${k_2}$ are the kernel scale. $\theta$ is the direction of the parallel strips. An effective adversarial example must satisfy both characteristics `without effecting human vision perception' and `attack ability'. (\emph{i}) Without effecting human vision perception refers to the difference between the adversarial example and the benign image must be small enough, i.e., $d{\rm{(}}{X_{adv}},X{\rm{)}} < \eta $. (\emph{ii})
Attack ability refers to the adversarial example generated by smoothing the linear texture of the benign image must be able to fool the target classifier, i.e., $f({X_{adv}},\tilde \vartheta ) = \hat y,\hat y \ne y$. Thus, to ensure that the adversarial examples generated by the Gabor filter do not affect human visual perception, we only smoothed features of different scales ${k_1}$ and angles $\theta$ in the benign image.
The set of adversarial examples generated by different $({k_1}, \theta)$ pairs, we call the initial adversarial example set. However, not all images generated by the Gabor filter can satisfy the characteristics of the adversarial example. Thus, we design a less perceptible-high attack success rate constraint to eliminate the unqualified examples from the initial example set.
\begin{table*}[t]
  \centering
  \caption{Accuracy of the threat model}
   \renewcommand\arraystretch{1.1}
  \setlength{\tabcolsep}{0.8mm}{
    \begin{tabular}{c|cccccccc}
    \toprule
    \multirow{2}[4]{*}{\diagbox{Method}{Dataset}} & \multicolumn{2}{c}{ResNext} & \multicolumn{2}{c}{Res2Net} & \multicolumn{2}{c}{Conformer} & \multicolumn{2}{c}{RepVGG} \\
\cline{2-9}    \multicolumn{1}{c|}{} & \multicolumn{1}{c}{Top-1} & \multicolumn{1}{c}{Top-5} & \multicolumn{1}{c}{Top-1} & \multicolumn{1}{c}{Top-5} & \multicolumn{1}{c}{Top-1} & \multicolumn{1}{c}{Top-5} & \multicolumn{1}{c}{Top-1} & \multicolumn{1}{c}{Top-5} \\
    \midrule
    Tiny-ImageNet & 78.88\% & 94.33\% & 79.19\% & 94.44\% & 83.82\% & 96.59\% & 81.81\% & 95.94\%\\
    CIFAR10 & 96.39\% & 99.95\% & 91.62\% & 99.62\% & 89.71\% & 99.61\% & 88.33\% & 99.45\% \\
    \midrule
    \multirow{2}[4]{*}{\diagbox{Method}{Dataset}} & \multicolumn{2}{c}{VGG16} & \multicolumn{2}{c}{ResNet50} & \multicolumn{2}{c}{MobileNet v2} & \multicolumn{2}{c}{SE-ResNet50} \\
\cline{2-9}    \multicolumn{1}{c|}{} & \multicolumn{1}{c}{Top-1} & \multicolumn{1}{c}{Top-5} & \multicolumn{1}{c}{Top-1} & \multicolumn{1}{c}{Top-5} & \multicolumn{1}{c}{Top-1} & \multicolumn{1}{c}{Top-5} & \multicolumn{1}{c}{Top-1} & \multicolumn{1}{c}{Top-5} \\
    \midrule
    Tiny-ImageNet & 71.62\% & 90.49\% & 76.55\% & 93.06\% & 71.86\% & 90.42\% & 77.74\% & 93.84\% \\
    CIFAR10 & 88.53\% & 99.71\% & 95.77\% & 99.10\% & 95.74\% & 99.93\% & 95.10\% & 99.89\% \\
     \bottomrule
    \end{tabular}}%
  \label{tab:addlabel}%
\end{table*}%
\begin{table*}[t]
  \centering
  \caption{Attack success rates of adversarial examples}
  \renewcommand\arraystretch{1}
  \setlength{\tabcolsep}{0.8mm}{
    \begin{tabular}{c|cccccccc}
    \toprule
    \multicolumn{9}{c}{CIFAR-10} \\
    \midrule
    \multicolumn{1}{c|}{\diagbox{Method}{Model}} & VGG   & Conformer & MobileNet & RepVGG & Res2Net & ResNet & ResNext & SE-ResNet \\
    \midrule
    DR    & 0.73  & 0.59  & 0.77  & 0.93  & 0.87  & 0.57  & 0.57  & 0.47 \\
    BIA   & 0.42  & 0.34  & 0.48  & 0.36  & 0.5   & 0.48  & 0.38  & 0.36 \\
    GUA   & 0.91  & 0.74  & 0.78  & 0.59  & 0.9   & 0.78  & 0.83  & 0.92 \\
    NPAttack & 0.4   & 0.2   & 0.4   & 0.2   & 0.2   & 0.2   & 0.2   & 0 \\
    AdvSmo   & \textbf{0.97}  & \textbf{0.81}  & \textbf{0.92}  & \textbf{0.94}  & \textbf{0.86}  &\textbf{ 0.83}  & \textbf{0.87}  & \textbf{1} \\
    \midrule
    \multicolumn{9}{c}{Tiny-ImageNet} \\
    \midrule
    \multicolumn{1}{c|}{\diagbox{Method}{Model}} & VGG   & Conformer & MobileNet & RepVGG & Res2Net & ResNet & ResNext & SE-ResNet \\
    \midrule
    DR    & 0.65  & 0.2   & 0.6   & 0.22  & 0.38  & 0.46  & 0.37  & 0.39 \\
    BIA   & 0.77  & 0.2   & 0.61  & 0.31  & 0.45  & 0.59  & 0.47  & 0.52 \\
    GUA   & 0.96  & 0.38  & 0.92  & 0.4   & 0.87  & 0.94  & 0.87  & 0.88 \\
    NPAttack & 0.2   & 0     & 0.5   & 0     & 0.1   & 0.3   & 0.4   & 0.2 \\
    AdvSmo   & \textbf{0.98}  & \textbf{0.85 } & \textbf{1}     & \textbf{0.88}  & \textbf{0.93}  & \textbf{0.98}  & \textbf{0.94}  & \textbf{0.93} \\
    \bottomrule
    \end{tabular}}%
  \label{tab:addlabel}%
\end{table*}%

\subsection{Less Perceptible-high Attack Success Rate Constraint}

Concerning the two characteristics of the adversarial examples, we determine the effective $({k_1}, \theta)$ pairs by resorting to several distance measurements. Our objective function is transformed into,
\begin{figure}[t]
    \centering
    \includegraphics[scale=0.3]{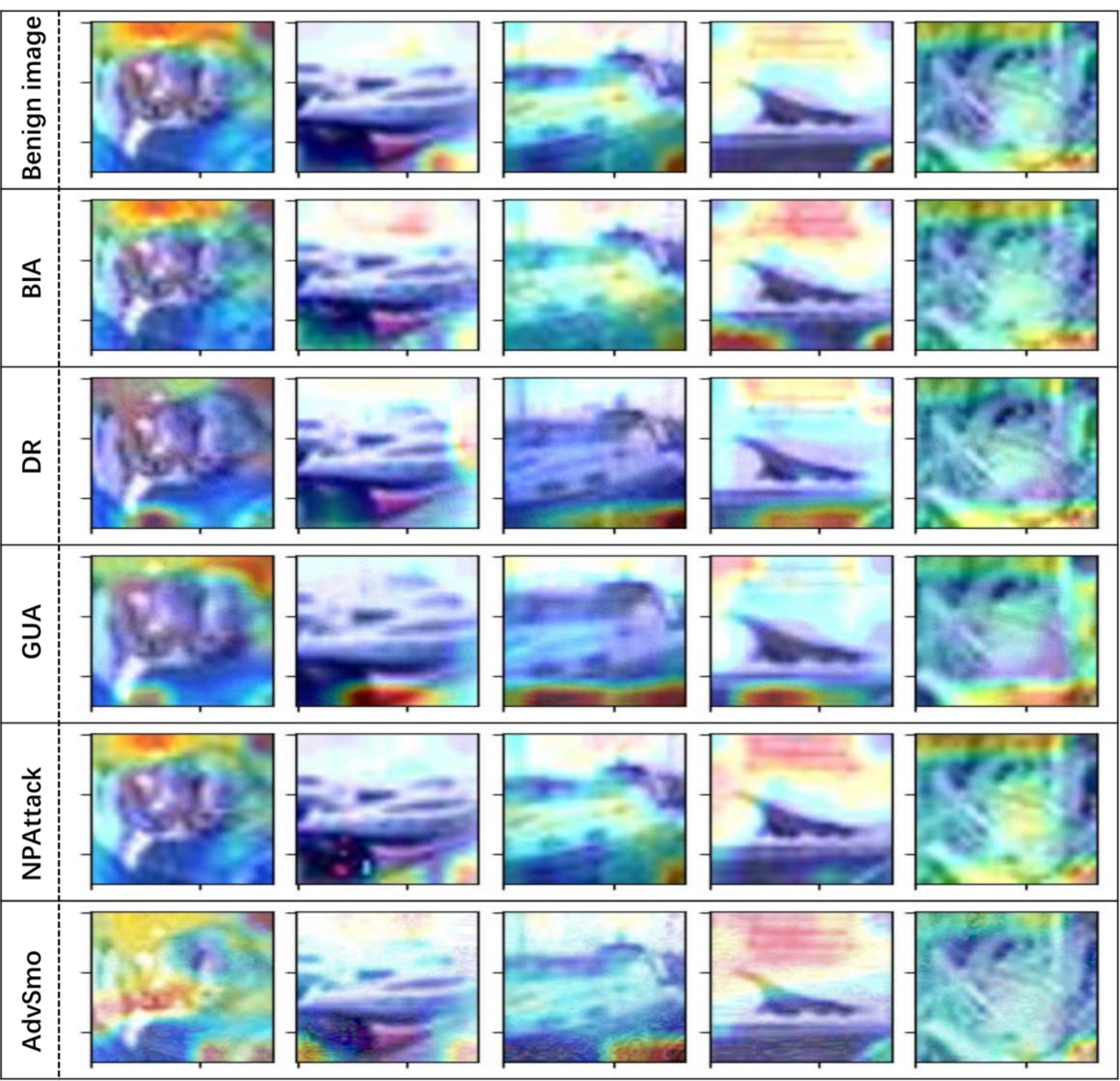}
    \caption{Grad-CAM}
\end{figure}
\begin{equation}
\begin{array}{l}
 \arg \mathop {\max }\limits_c f({X_{adv}},\tilde \vartheta ) = \hat y \\
 s.t.,{\rm{ }}{d_1}({X_{adv}},X{\rm{)}} < {\eta _1}, \cdots ,{d_n}({X_{adv}},X) < {\eta _n},\hat y \ne y, \\
 \end{array}
\end{equation}

${\eta _1},{\eta _2}, \cdots ,{\eta _n}$ is the thresholds of different distance measurements. According to Eq. (4), the valid $({k_1}, \theta)$ pairs can be determined as,

\begin{equation}
U = {U_{{d_1}}} \cap {U_{{d_1}}} \cap , \cdots ,{U_{{d_n}}}.
\end{equation}
The distance measurements used to determine the $({k_1}, \theta)$ pairs are not chosen arbitrarily but are based on the properties of the Gabor filter processing on generated images. We find that different $({k_1}, \theta)$ pairs of generated images differ significantly in terms of image luminance, contrast, and structural similarity. Thus we screen $({k_1}, \theta)$ pairs based on these three metrics. The Structure Similarity Index Measure (SSIM) can meet our requirements.

Using the SSIM to eliminate invalid $({k_1}, \theta)$, we are surprised to find that SSIM significantly reduce computerization. The SSIM values of the new image constructed by $({k_1}, \theta)$ are symmetric about 90 degrees, the range of ${k_1}$ can be reduced to [0, 90].

Although SSIM can remove some of the invalid $({k_1}, \theta)$ pairs, there are anomalous points (low attack success rate or too different from benign images). To eliminate the $({k_1}, \theta)$ pairs with a low attack success rate, we use the Mean Square Error (MSE) to eliminate the $({k_1}, \theta)$ pairs with a low attack success rate (less than 80\%).
To eliminate the images that differ too much from the benign image, we use the ${L_\infty }$ norm to discard the invalid adversarial examples. Thus,

\begin{equation}
\begin{array}{l}
 \arg \mathop {\max }\limits_c f({X_{adv}},\tilde \vartheta ) = \hat y, \\
 s.t., \\
 {{\eta _{11}} < \rm{ }}SSIM({X_{adv}},X{\rm{)}} < {\eta _{12}}, \\
 {\rm{      }}{\eta _{21}} < MSE({X_{adv}},X) < {\eta _{22}}, \\
 {\rm{     }}{\eta _{31}} < |{X_{adv}} - X{|_\infty } < {\eta _{32}}, \\
 {\rm{      }}\hat y \ne y. \\
 \end{array}
\end{equation}

The image generated ${X^{'}}$ by the Gabor filter is expressed as ${X^{'}} = G(k,\theta )$. The $({k_1}, \theta)$ pairs determined by SSIM is ${U_{SSIM}} = \{ {k_1},\theta |SSIM(G({k_1},\theta ),X{\rm{)}} < {\eta _1}\}$, by the MSE is  ${U_{MSE}} = \{ {k_1},\theta |MSE(G({k_1},\theta ),X{\rm{)}} < {\eta _2}\}$, by ${L_\infty }$ is ${U_{{L_\infty }}} = \{{k_1},\theta ||G({k_1},\theta ),X{|_\infty } < {\eta _3}\}$. Taking the intersection of the above three sets, we get the final set U of $({k_1}, \theta)$ pairs.
\begin{equation}
\begin{array}{l}
 U = {U_{SSIM}} \cap {U_{MSE}} \cap {U_{{L_\infty }}}
 \end{array}.
\end{equation}

To determine the value of ${\eta _{11}}$ and ${\eta _{12}}$, we design a DNN to fit the relationship between $({k_1}, \theta)$ and SSIM. The structure of our network is a fully connected neural network with 1 inputs $({k_1}, \theta)$, 1 output (the value of SSIM), and 10 neurons. It is important to note that $({k_1}, \theta)$ needs to be normalized before feeding it to the DNN, the optimizer is Adam optimizer, the training period epoch is set to 700, and the loss function is MSE. ${\eta _{21}}$ and  ${\eta _{22}}$ depends on what kind of attack success rate we expect to generate. We regard $({k_1}, \theta)$ pairs with an attack success rate below 80\% as invalid. ${\eta _{31}}$ and  ${\eta _{32}}$ is determined by interviewing 200 students. We set ${\eta _{11}}=0.077444225$, ${\eta _{12}}=0.132868965$, ${\eta _{21}}=0.020213895$, ${\eta _{22}}=0.038001586$, ${\eta _{31}}=19.81960784$, and ${\eta _{32}}=27.19215686$.

\begin{table*}[t]
  \centering
  \caption{Comparison of average attack success rate}
  \renewcommand\arraystretch{1}
  \setlength{\tabcolsep}{0.8mm}{
    \begin{tabular}{c|cccc|c}
    \toprule
    \diagbox{Method}{Model}   & DR    & BIA    & GUA   & NPAttack    & AdvSmo \\
    \midrule
    CIFAR-10 & 0.6875 & 0.415 & 0.80625 & 0.225 & \textbf{0.9} \\
    Tiny-ImageNet & 0.40875 & 0.49  & 0.7775 & 0.5875 & \textbf{0.93625} \\
    \midrule
    Average & 0.548125 & 0.4525 & 0.791875 & 0.40625 & \textbf{0.918125} \\
    \bottomrule
    \end{tabular}}%
  \label{tab:addlabel}%
\end{table*}%
\section{Experiment Results}

\subsection{Experimental Setup}
To verify the effectiveness of AdvSmo, the eight threat models are ResNext \cite{xie2017aggregated}, Res2Net \cite{gao2019res2net}, Conformer \cite{peng2021conformer}, RepVGG \cite{ding2021repvgg}, VGG16 \cite{simonyan2014very}(The target model to determine ${\eta}$), ResNet50 (i.e., ResNet)\cite{he2016deep}, MobileNet v2 (i.e., MobileNet) \cite{howard2017mobilenets}, and SE-ResNet50 (i.e., SE-ResNet) \cite{hu2018squeeze}, the four compared black-box attack schemes are BIA \cite{zhang2022beyond}, DR \cite{lu2020enhancing}, GUA \cite{poursaeed2018generative}, and NPAttack \cite{bai2020improving},  the six adversarial defensive schemes are Bilinear Filtering, Gaussian Filtering, Max Filtering, Mean Filtering, Median Filtering, and Min Filtering, and the datasets are CIFAR-10 and Tiny-ImageNet. The accuracy of threat models is shown in Table 1.

\subsection{High Attack Success Rate and Strong Transferability Validation}
Table 2 shows the attack success rates of adversarial examples generated by the five black-box attack schemes, BIA \cite{zhang2022beyond}, DR \cite{lu2020enhancing}, GUA \cite{poursaeed2018generative}, NPAttack \cite{bai2020improving}, and AdvSmo, against eight threat models. This table shows that the success rate of AdvSmo against eight target models is significantly better than the other four attack schemes, and the success rate of AdvSmo against VGG16 on the Tiny-ImageNet dataset can reach 98\%. Table 3 shows the average attack success rates of adversarial examples generated by the five schemes BIA, DR, GUA, NPAttack, and AdvSmo against eight target models on the CIFAR 10 dataset and the Tiny-ImageNet dataset. This table shows that the average attack success rate of AdvSmo increased by 37\% compared to DR, 47\% compared to BIA, 13\% compared to GUA, and 51\% compared to NPAttack. Based on the above analysis, it is clear that the adversarial examples generated by AdvSmo have a high attack success rate and are highly transferable.

\subsection{Neural Network Attention Analysis}
Figure 2 visualizes the texture by the GLCM for the adversarial examples generated by the five attack methods. Red markers indicate areas of texture change. From this figure, it can be seen that five black-box attack methods affect the texture features of the images to different degrees when generating adversarial examples. This figure shows that NPAttack has the most negligible impact on the texture features of the image while AdvSmo has the greatest impact on the image texture features. Analyze together with Tables 2-3 and this figure, we find that the degree of the image texture information changes determines the attack success rate of the generated adversarial examples, i.e., the more the image texture features are modified, the higher the attack success rate and the stronger the transferability of the adversarial examples.

To better demonstrate the intrinsic principle of AdvSmo, we visualize the Grad-CAM \cite{selvaraju2017grad} of the adversarial example, as shown in Figure 3. From this figure, the feature mapping of the adversarial examples generated by AdvSmo has been artificially modified compared to the feature mapping of benign images. Compared with the other four adversarial attack schemes, the feature mapping of the adversarial example generated by AdvSmo is changed to the greatest extent. Thus, AdvSmo is able to mislead the target model to misclassify.

\section{Conclusions}
Designing effective adversarial attack methods helps to inspire more robust defense schemes, but the adversarial examples generated by previous adversarial attack methods mainly face two problems: poor transferability and the inability to evade adversarial defense mechanisms. We propose a black-box adversarial attack by smoothing the linear structure of texture. The detailed experimental results show that the adversarial examples generated by AdvSmo have a high attack success rate, strong transferability, and strong evasion. In future research, we will continue to explore novel adversarial example generation methods.

\bibliographystyle{alpha}
\bibliography{neurips_2022-new}
\end{document}